\documentclass[lettersize,journal]{IEEEtran}
\usepackage{amsmath,amsfonts}
\usepackage{algorithmic}
\usepackage{algorithm}
\usepackage{array}
\usepackage[caption=false,font=normalsize,labelfont=sf,textfont=sf]{subfig}
\usepackage{textcomp}
\usepackage{stfloats}
\usepackage{url}
\usepackage{verbatim}
\usepackage{graphicx}
\usepackage{cite}
\usepackage{algorithmic}
\usepackage{lipsum}

\usepackage{cuted}
\usepackage{wrapfig}
\usepackage{newfloat}
\usepackage{hhline}
\usepackage{listings}
\usepackage{tikz}
\usepackage{latexsym}
\usepackage[T1]{fontenc}
\usepackage[utf8]{inputenc}
\usepackage{microtype}
\usepackage{inconsolata}
\usepackage{graphicx}
\usepackage{multirow}
\usepackage{algorithmic}
\usepackage{makecell}
\usepackage{algorithm}
\usepackage{amsmath,amsfonts}
\usepackage{textcomp}
\usepackage{url}
\usepackage{array}
\usepackage{amssymb}
\usepackage{booktabs}
\usepackage{xcolor}  
\usepackage{color}
\usepackage{colortbl}
\definecolor{ljcolor}{RGB}{34,139,34}
\usepackage{pifont}
\definecolor{backgroud}{RGB}{240,240,240}
\definecolor{hdash}{RGB}{191, 191, 191}
\definecolor{ent1}{RGB}{155,180,227}
\definecolor{ent2}{RGB}{239,148,158}
\definecolor{ent3}{RGB}{125,223,215}
\definecolor{ent4}{RGB}{254,217,97}
\definecolor{ent5}{RGB}{169,131,198}
\definecolor{ent6}{RGB}{172,215,142}
\definecolor{ent7}{RGB}{37,227,255}
\definecolor{ent8}{RGB}{0,0,255}
\definecolor{ent9}{RGB}{191,191,191}
\definecolor{ent10}{RGB}{181,139,1}
\definecolor{ent11}{RGB}{0,255,0}
\definecolor{ent12}{RGB}{36,144,135}
\definecolor{ent13}{RGB}{200,29,49}
\definecolor{ent14}{RGB}{249,43,237}
\definecolor{ent15}{RGB}{255,0,0}
\definecolor{uc}{RGB}{77, 191, 249}
\definecolor{bc}{RGB}{5,178,83}
\definecolor{hdash}{RGB}{191, 191, 191}
\usepackage{hyperref}
\hypersetup{
    colorlinks=true,
    citecolor=blue,
    linkcolor=blue, 
    urlcolor=blue 
}
\newcommand{\mysize}{\fontsize{5pt}{4pt}\selectfont}
\usepackage{arydshln}
\usepackage[most]{tcolorbox}
\usepackage{bibentry}

\usepackage{CJKutf8}

\begin{document}

\title{{M$^{3}$D}: A \underline{M}ultimodal, \underline{M}ultilingual and \underline{M}ultitask \underline{D}ataset for Grounded Document-level Information Extraction}

\author{Jiang Liu, Bobo Li, Xinran Yang, Na Yang, Hao Fei, Mingyao Zhang, Fei Li*, Donghong Ji*
\thanks{Jiang Liu, Bobo Li, Xinran Yang, Donghong Ji are with the Key Laboratory of Aerospace Information Security and Trusted Computing, Ministry of Education, School of Cyber Science and Engineering, Wuhan University (e-mail: liujiang@whu.edu.cn; boboli@whu.edu.cn; 2021302191942@whu.edu.cn; dhji@whu.edu.cn).}
\thanks{Fei Li is with the Key Laboratory of Aerospace Information Security and Trusted Computing, Ministry of Education, School of Cyber Science and Engineering, Wuhan University, and Laboratory for Advanced Computing and Intelligence Engineering, Wuxi, China (e-mail: lifei\_csnlp@whu.edu.cn).}
\thanks{Na Yang, Mingyao Zhang are with the School of Foreign Languages and Literature, Wuhan University (e-mail: 2022201020044@whu.edu.cn; myzhang@whu.edu.cn).}
\thanks{Hao Fei is with the School of Computing, National University of Singapore (e-mail: haofei37@nus.edu.sg).}
\thanks{Donghong Ji is the first corresponding author.}
\thanks{Fei Li is the second corresponding author.}
}


\maketitle

\begin{abstract}
Multimodal information extraction (IE) tasks have attracted increasing attention because many studies have shown that multimodal information benefits text information extraction. However, existing multimodal IE datasets mainly focus on sentence-level image-facilitated IE in English text, and pay little attention to video-based multimodal IE and fine-grained visual grounding.
Therefore, in order to promote the development of multimodal IE, we constructed a multimodal multilingual multitask dataset, named M$^{3}$D, which has the following features:
(1) It contains paired document-level text and video to enrich multimodal information;
(2) It supports two widely-used languages, namely English and Chinese;
(3) It includes more multimodal IE tasks such as 
entity recognition, entity chain extraction, relation extraction and visual grounding. 
In addition, our dataset introduces an unexplored theme, i.e., biography, enriching the domains of multimodal IE resources.
To establish a benchmark for our dataset, we propose an innovative hierarchical multimodal IE model. This model effectively leverages and integrates multimodal information through a Denoised Feature Fusion Module (DFFM). Furthermore, in non-ideal scenarios, modal information is often incomplete. Thus, we designed a Missing Modality Construction Module (MMCM) to alleviate the issues caused by missing modalities.
Our model achieved an average performance of 53.80\% and 53.77\% on four tasks in English and Chinese datasets, respectively, which set a reasonable standard for subsequent research. In addition, we conducted more analytical experiments to verify the effectiveness of our proposed module. We believe that our work can promote the development of the field of multimodal IE.
\end{abstract}


\section{Introduction}
Information extraction (IE) aims to identify and extract pre-defined information from unstructured text. 
IE is a hot topic in the natural language processing (NLP) community, which usually includes the following widely-studied tasks, such as named entity recognition (NER) \cite{introduction_ner}, coreference resolution (CR) \cite{introduction_EL}, relation extraction (RE) \cite{introduction_RE} and event extraction (EE) \cite{introduction_EE}.
Various methods have been investigated for solving these tasks.
For example, sequence-to-sequence (Seq2Seq) methods \cite{bourd_error_1} solve NER by directly generating entities, while span-based methods \cite{span_ner} extract entities by enumerating spans and grid-tagging-based methods \cite{tagging_ner} convert sentences into two-dimensional (2D) table for entity extraction. 
To solve CR and RE,
some studies first extract entities and then identify coreference relations or entity relations \cite{pipline_EL,pipline_RE},
but they inevitably produce error propagation. Therefore, more advanced methods directly generates entity chains or relation pairs by implementing end-to-end models \cite{end2end_EL,end2end_RE}.

\begin{figure}[!t]
\centering
\includegraphics[width=0.5\textwidth]{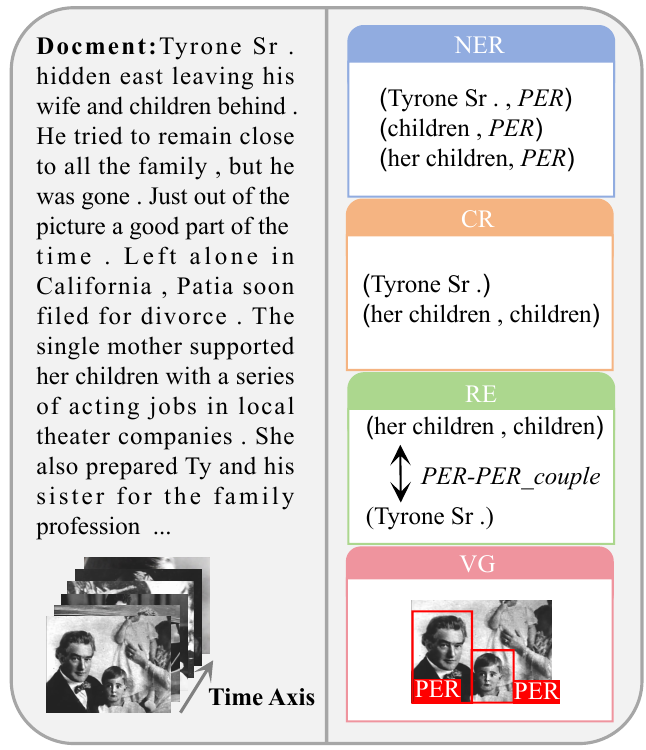}
\caption{
A sample in the M$^{3}$D dataset.
The left part is the input example, and the right part is the output example of four tasks.
} 
\label{fig1}
\end{figure}

Despite achieving certain success, existing studies only make use of text information. 
In addition to a large amount of text information, the Internet also contains a large amount of image, audio and video information. Therefore, Lu et al. \cite{tweet-2015}, Zhang et al. \cite{tweet-2017} and Sui et al. \cite{audio_ner} respectively proposed multimodal NER (MNER) datasets containing image information. 
Yu et al. \cite{gmner} further improved their work by proposing a grounded MNER dataset that also focuses on the recognition of visual information. 
Besides MNER, multimodal CR (MCR) \cite{image_cr} and multimodal RE (MRE) \cite{mre,re_video_data} have also been investigated with curated datasets.
Based on these multimodal datasets, researchers have also conducted a large number of studies demonstrating that combining multiple modal information can facilitate these information extraction tasks \cite{introduction_multi_ner,introduction_multi_cr,introduction_multi_re}. 


However, these studies still have some shortcomings. As shown in Table \ref{tab:dataset_compar}, Twitter-2015 \cite{tweet-2015} and Twitter-2017 \cite{tweet-2017} only contain image but ignore video.
Even if MDocRE \cite{re_video_data} contains video information, it neglects the extraction of visual objects.
Some datasets such as MNRE \cite{related_mre} focus on sentence-level information extraction and have not considered complex document-level tasks.
Although DocRED \cite{docred} is a document-level RE dataset, it does not leverage multimodal information.
Last but not least, most datasets contains single language \cite{image_cr,mre} or limited information extraction tasks \cite{tweet-2015,audio_ner}.

Considering the shortcomings in previous work, we constructed a \textbf{M}ultimodal, \textbf{M}ultilingual and \textbf{M}ultitask \textbf{D}ataset (M$^{3}$D) for grounded
document-level information extraction.
We first crawled English videos and Chinese videos from the English website YouTube and the Chinese website bilibili respectively, then we split the videos into video clips, and then used the subtitle generation tool to generate subtitles as text information, thus obtaining 4093 multimodal samples. Finally, we manually annotated these samples, including entities, entity chains, entity relations, and visual targets. 
A sample in our M$^{3}$D dataset is shown in Figure \ref{fig1}. The left part represents the input text and video, with the video being equivalent to a sequence of images containing temporal information. The right part represents the output for the four tasks, which includes manually annotated entities, coreference chains, relations and grounded bounding boxes. The comparison between M$^{3}$D and other datasets is shown in Table \ref{tab:dataset_compar}. 
\begin{table*}[!t]
\fontsize{8}{10}\selectfont
\setlength{\tabcolsep}{0.8mm}
\centering
\caption{Comparison between M$^{3}$D and other datasets. ``—'' indicates that there are no reports in the original papers. (Doc.: Document, Sent.: Sentence, Ent.: Entity, Rel.: Relation, Cha.: Entity Chain, Gro.: Visual Grounding)}
\resizebox{1\textwidth}{!}{
\begin{tabular}{l c c c c c c c c c c c c c c c c c c}
\hline
\bf Dataset & \phantom{}&  \bf \# Doc. &\phantom{}& \bf \# Word &\phantom{}& \bf \# Sent. &\phantom{}& \bf \# Ent. (Type) &\phantom{} & \bf \# Rel. (Type) &\phantom{}& \bf \# Cha. &\phantom{}& \bf \# Gro. &\phantom{}& \bf Multilingual &\phantom{}& \bf Multimodal
\\
\hline
Twitter-2015&\phantom{}&—&\phantom{}&133k&\phantom{}&8,257&\phantom{}&13,168 (4)&\phantom{}&—&\phantom{}&—&\phantom{}&—&\phantom{}&\ding{55}&\phantom{}&image\\
Twitter-2017&\phantom{}&—&\phantom{}&76k&\phantom{}&4,819&\phantom{}&8,724 (4)&\phantom{}&	—&\phantom{}&—&\phantom{}&—&\phantom{}&	\ding{55}&\phantom{}&	image\\
MNRE	&\phantom{}&—&\phantom{}&	258k&\phantom{}&	9,201&\phantom{}&	30,970 (4)&\phantom{}&	5,767 (23)&\phantom{}&	—&\phantom{}&—&\phantom{}&	\ding{55}&\phantom{}&	image\\
DocRED&\phantom{}&	5,053&\phantom{}&	1,002k&\phantom{}&	40,276&\phantom{}&	132,375 (6)&\phantom{}&	63,427 (96)&\phantom{}&	98,560&\phantom{}&	—&\phantom{}&\ding{55}&\phantom{}&	—\\
MDocRE	&\phantom{}&3,043&\phantom{}&	1,372k&\phantom{}&	70,703&\phantom{}&		— (4)&\phantom{}& — (21)		&\phantom{}&—&\phantom{}&—&\phantom{}&	\ding{51}&\phantom{}&	video\\
\hdashline 
M$^{3}$D&\phantom{}&	4,093&\phantom{}&	1,186k&\phantom{}&	50,271&\phantom{}&	62,343 (4)&\phantom{}&	24,568 (34)&\phantom{}&	33,095&\phantom{}&16,891&\phantom{}&	\ding{51}&\phantom{}&	video\\
\hline
\end{tabular}
}
\label{tab:dataset_compar}
\end{table*}


To establish a benchmark for our dataset, we designed a hierarchical multimodal information extraction model. Considering the inadequate utilization and fusion of multimodal information in existing models, we developed a denoised feature fusion module (\textbf{DFFM}). Specifically, this module first employs a variational auto-encoder (VAE) to denoise features from different layers of a pre-trained model, maximizing the utilization of effective modal information. The multilayer features from the pre-trained model are then categorized into low-level, mid-level, and high-level features, which are fused with other modal information to ensure comprehensive integration of effective modalities.
Furthermore, under non-ideal conditions, modal information is often incomplete, which can lead to a decrease in the robustness and performance of the model. To alleviate this issue, we designed a missing modality construction module (\textbf{MMCM}). Specifically, we concatenate the available modal information with the missing modality prompt features, which are then processed through convolutional layers to generate information for the missing modality. We compared with seven baseline models, including two large language models (LLMs), and our method achieved the best average performance on four tasks, reaching 53.80\% and 53.77\% on English and Chinese datasets respectively. We also conducted various analytical experiments such as ablation experiments to further demonstrate the effectiveness of our proposed module. The contributions of this paper include:

$\bullet$ We proposed a multimodal, multilingual, and multitask information extraction dataset to address the shortcomings of the existing datasets.

$\bullet$ We have designed an innovative hierarchical model to address all multimodal IE tasks and established a reasonable baseline for subsequent research. Specifically, we developed two modules for this model. The first is a denoised feature fusion module that enables the model to fully utilize and integrate multimodal information. The second is a missing modality construction module that alleviates the issues caused by missing modalities for the model.

$\bullet$ Our experiments demonstrate that our model performs well on multimodal information extraction. In addition, we conducted more analytical experiments to verify the effectiveness of our proposed module. Our code will be released later.


\section{Related Work}
\subsection{Multimodal Named Entity Recognition} 

Lu et al. \cite{tweet-2015} first conducted research on MNER and proposed the first MNER dataset. Zhang et al. \cite{tweet-2017} introduced an adaptive joint attention network to fuse multimodal information and constructed a MNER dataset for evaluation. Yu et al. \cite{related_work_ner_1} first paid attention to the alignment problem of images and texts, and designed a multimodal interaction module to obtain image-aware word representation and word-aware visual representation.  Zhou et al. \cite{related_work_ner_semi_supervised} proposed a novel span-based multimodal variational autoencoder (SMVAE) model for semi-supervised MNER. Zhao et al. \cite{related_work_ner_entity_level} ensured text and image alignment by building heterogeneous graphs to capture visual information related to entities. Li et al. \cite{related_work_ner_prompt} promoted the recognition ability of MNER models by introducing external knowledge.

\subsection{Multimodal Coreference Resolution}
Goel et al. \cite{image_cr} first studied the MCR task and proposed a MCR dataset. Guo
et al. \cite{related_work_cr} combined graph neural networks with VL-BERT to solve the coreference problem of multi-turn dialogues containing scene graphs. Goel et al. \cite{introduction_multi_cr} believed that fine-grained image text alignment methods are difficult to perform document-level tasks, so they proposed a data efficient semi-supervised approach that utilizes image-narration pairs to resolve coreferences and narrative grounding in a multimodal context.

\subsection{Multimodal Relation Extraction}
Zheng et al. \cite{mre} first studied the MRE task and proposed a MRE dataset. Zheng
et al. \cite{related_mre} developed a dual graph alignment method to capture the correspondence between images and text.  Chen
et al. \cite{related_prefix} constructed hierarchical multi-scaled visual features through a dynamic gated aggregation strategy and integrated them into the model as prefix prompts.  Zheng et al. \cite{related_Translation} believed that the inconsistency between modalities is similar to the cross-language difference problem, so they regarded images and texts as translations of each other to enhance the alignment between images and texts. Wu et al. \cite{related_Information} constructed a cross-modal graph (CMG) to solve the challenges of internal-information over-utilization and external-information under-exploitation in MRE. Yuan et al. \cite{introduction_multi_re} proposed an edge-enhanced graph alignment network and a word-pair relation tagging (EEGA) to consider the interaction between MNER and MRE to jointly solve the two tasks.

\subsection{Visual Grounding}
Visual Grounding (VG) aims to locate the most relevant visual regions from images through entity descriptions in text. Existing work on VG is mostly divided into two categories. The first is to directly predict visual regions \cite{end2endvg_1,end2endvg_2} using object detection models such as YOLO \cite{yolo} or SSD \cite{ssd}. The second is to first use the object detection model to obtain region suggestions, and then rank them according to their relevance to the text query \cite{twovg,introduction_multi_ner}. 
\begin{figure*}[!t]
\centering
\includegraphics[width=0.9\textwidth]{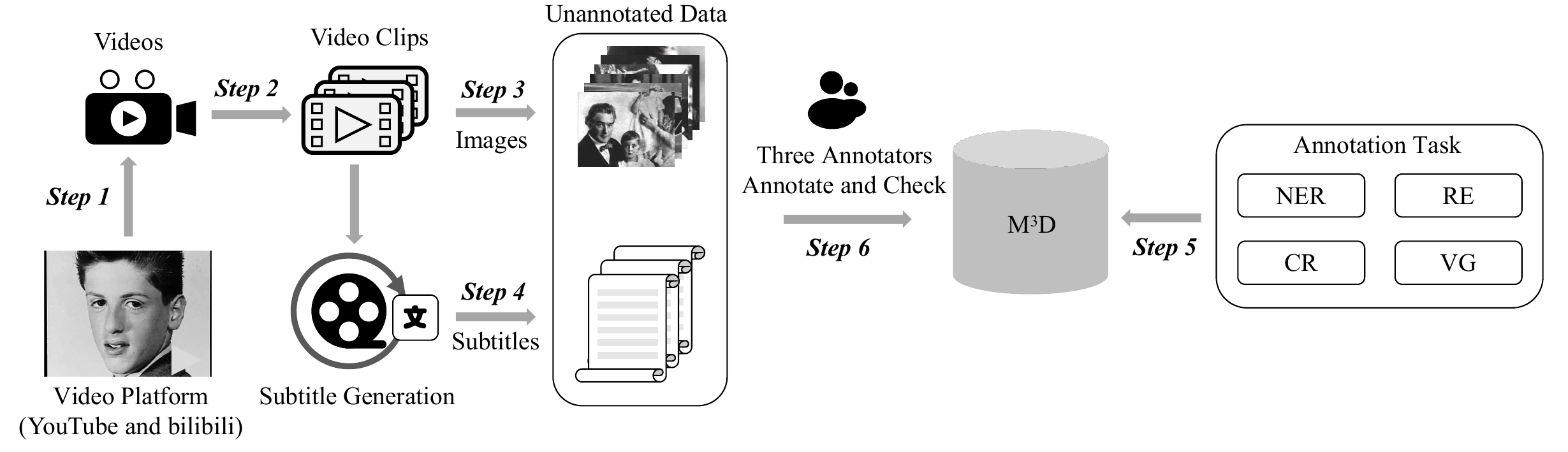}
\caption{The overall construction process of the M$^3$D dataset. \textbf{Step 1:} Crawl videos from video platforms. \textbf{Step 2:} The video is split into video clips. \textbf{Step 3:} Video clips are sampled as images. \textbf{Step 4:} Generate subtitles from video clips using a subtitle generation model. \textbf{Step 5:} Develop annotation guidelines to guide annotations. \textbf{Step 6:} Three annotators annotate image and text data based on annotation guidelines. (NER: named entity recognition, RE: relation extraction, CR: coreference resolution, VG: visual grounding)
}
\label{Data_Construction}
\end{figure*}

\subsection{Missing Modality}
Tran et al. \cite{missing_work_1} first paid attention to the problem of missing modalities in target recognition. They used cascaded residual autoencoder (CRA) to fill in the missing modalities. Wang et al. \cite{missing_drop} proposed a modality dropout (m-drop) and a multimodal sequential autoencoder (m-auto) to learn multimodal representations to alleviate the problem of missing modalities in recommendation systems. Cho et al. \cite{missing_know_visual} and Li et al. \cite{missing_know_msa} used knowledge distillation to deal with the missing modalities in visual question answer-difference prediction task and multimodal sentiment analysis (MSA). Zhao et al. \cite{missing_mmin} proposed a missing modality imagination network (MMIN) to deal with the missing modalities in MSA. Zeng et al. \cite{missing_zeng} proposed an ensemble-based missing modality reconstruction (EMMR) network to detect and restore the semantic features of key missing modalities to solve the modality problem in MSA. Lee et al. \cite{missing_prompt_vr} and Guo et al. \cite{missing_prompt_msa} constructed missing modalities through prompts to solve the problem of missing modalities in visual recognition and MSA.



\section{Dataset Construction}
We construct a new dataset to facilitate multimodal tasks. The overall process of building the M$^3$D dataset is shown in Figure \ref{Data_Construction}. Our original English and Chinese videos come from YouTube\footnote{\href{https://www.youtube.com}{https://www.youtube.com}} and bilibili\footnote{\href{https://www.bilibili.com}{https://www.bilibili.com}} respectively.

\textbf{\textit{Step 1:}} The types of videos we crawled from both websites were biographical videos. The main reason is that such videos will mention entities such as person and location many times, and when a specific entity is mentioned, corresponding images are usually shown in the video, allowing for meaningful supplementation of information between the two modalities. 

\textbf{\textit{Step 2:}} Since these videos are usually longer, with most videos lasting more than 30 minutes, we will split the video into multiple video clips and keep them around one to two minutes long. We split the video based on the length of the subtitles. We then obtain unannotated image and text data based on these video clips.

\textbf{\textit{Step 3:}} We sample each video clip, we sample an image every 24 frames, and finally get the image data to be annotated.

\textbf{\textit{Step 4:}} In order to ensure that the video is aligned with the text, we use the subtitle generation tool\footnote{\href{https://github.com/YaoFANGUK/video-subtitle-generator}{https://github.com/YaoFANGUK/video-subtitle-generator}} to generate subtitles and check and modify errors, and then use them as text data to be annotated. Therefore, our English data and Chinese data are non-aligned. There are three reasons why we did not align the Chinese and English datasets: 1) The amount of data in some small languages is usually very limited, and translation alignment is required, while both Chinese and English have sufficient resources; 2) Proper nouns such as names of people or places are difficult to align into appropriate Chinese; 3) We believe that this can maintain the diversity of the data.

\textbf{\textit{Step 5:}} We have developed annotation tasks and corresponding rules to guide annotators in annotation. Please refer to Section \ref{Annotation Specifications} for detailed annotation specifications.

\textbf{\textit{Step 6:}} Three annotators manually annotated the English and Chinese data. The BRAT text annotation tool\footnote{\href{https://brat.nlplab.org/index.html}{https://brat.nlplab.org/index.html}} was used for text data annotation, and the LabelImg image annotation tool\footnote{\href{https://github.com/HumanSignal/labelImg}{https://github.com/HumanSignal/labelImg}} was used for image data annotation.
The three annotators were divided into two groups to annotate the same data. If there was any inconsistency, the three annotators would discuss and decide. The final Cohen's Kappa score was 78.18\%, which shows that our data achieved high consistency.

\begin{table}[!t]
\fontsize{5}{7}\selectfont
\setlength{\tabcolsep}{0.8mm}
\centering
\caption{Statistics for our dataset.}
\resizebox{0.4\textwidth}{!}{
\begin{tabular}{l l c c c c c c}
\hline
&&\phantom{}&\bf Train&\phantom{}&\bf Dev&\phantom{}&\bf Test\\
\hline
\multirow{5}{*}{\bf EN} & \# \bf Doc. & \phantom{} & 1,644 &\phantom{} & 205 & \phantom{} & 207\\
 & \# \bf Ent. & \phantom{} & 27,843 &\phantom{} & 3,457 & \phantom{} & 3,587\\
 & \# \bf Rel. & \phantom{} & 11,410 &\phantom{} & 1,388 & \phantom{} & 1,364\\
 & \# \bf Cha. & \phantom{} & 15,031 &\phantom{} & 1,885 & \phantom{} & 1,882\\
 & \# \bf Gro. & \phantom{} & 9,812 &\phantom{} & 1,171 & \phantom{} & 1,198\\
\hdashline
\multirow{5}{*}{\bf ZH} & \# \bf Doc. & \phantom{} & 1,629 &\phantom{} & 203 & \phantom{} & 205\\
 & \# \bf Ent. & \phantom{} &22,043 &\phantom{} & 2,685 & \phantom{} & 2,728\\
 & \# \bf Rel. & \phantom{} &8,372 &\phantom{} &985 & \phantom{} & 1,049\\
 & \# \bf Cha. & \phantom{} &11,481 &\phantom{} & 1,397 & \phantom{} & 1,419\\
 & \# \bf Gro. & \phantom{} & 3,726 &\phantom{} & 505 & \phantom{} & 482\\
\hline
\end{tabular}
}
\label{tab:dataset_Statistics}
\end{table}

\begin{figure}[!t]
\centering
\includegraphics[width=0.45\textwidth]{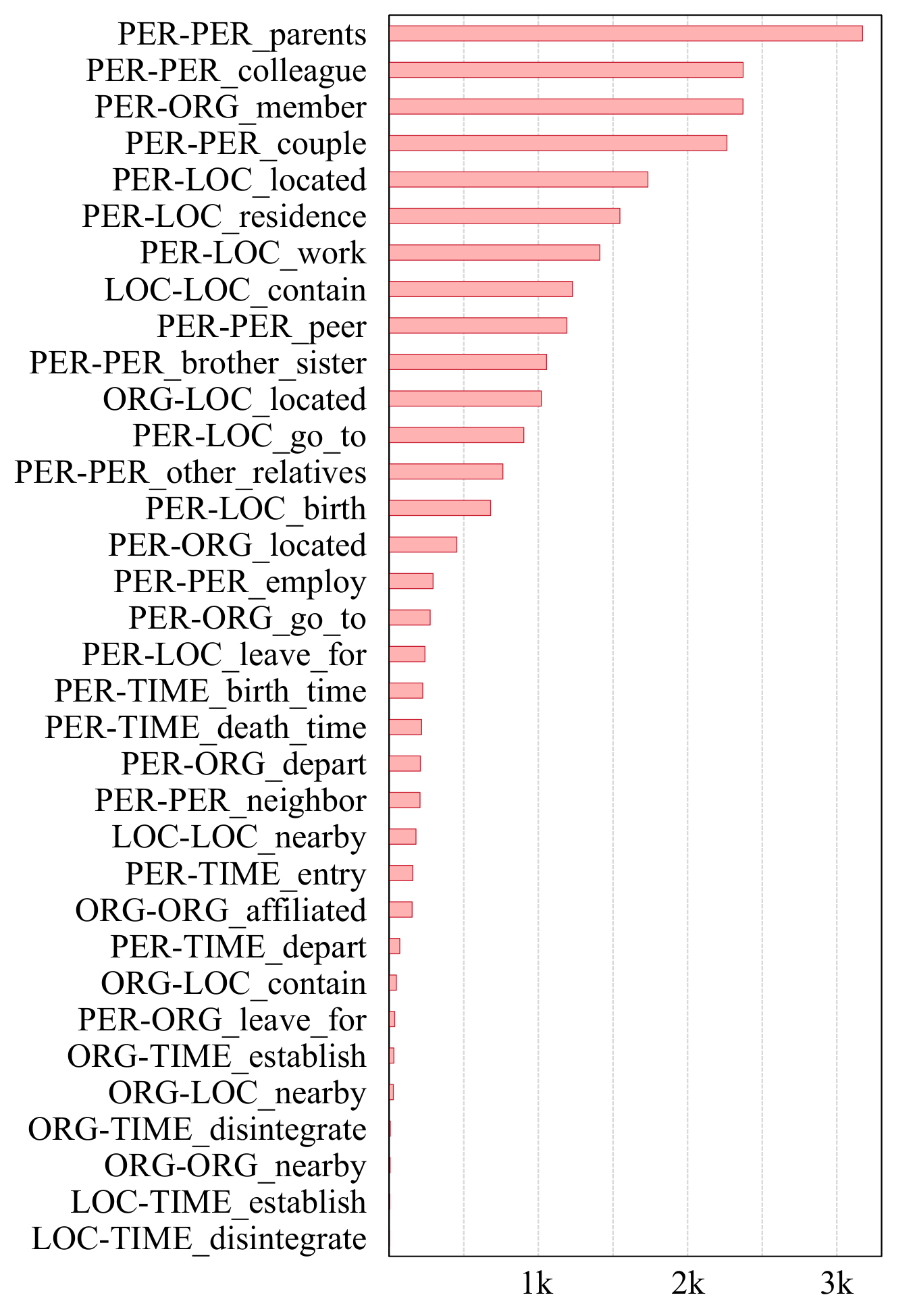}
\caption{Quantity statistics for each relation type.  } 
\label{Quantity_relation}
\end{figure}

\textbf{Dataset Statistics} Our dataset contains 4 entity types, 34 relation types and 3 visual object types. The entity types are PER, LOC, ORG and TIME respectively, and the corresponding statistical numbers are 42,620, 10,137, 5,462 and 4,124. The visual object types are PER, LOC and ORG respectively, and the corresponding statistical numbers are 15,599, 845 and 447. The detailed statistical results of relation types are shown in Figure \ref{Quantity_relation}. Then we split the annotated English and Chinese datasets into training sets, development sets and testing sets at an 8:1:1 ratio. The specific statistics of our dataset are summarized in Table \ref{tab:dataset_Statistics}. 

\section{Annotation Specifications}
\label{Annotation Specifications}
\subsection{Entity Annotation Specification}
For entity recognition tasks, we only label specific entities and general entities, and do not label entities that are only pronouns. An example is shown below:
\begin{tcolorbox}
\textcolor{ent8}{Tyrone Sr .} hidden east leaving \textcolor{ent13}{his wife} and \textcolor{ent13}{children} behind . \textcolor{ent4}{He} tried to remain close to all the family , but he was gone . Just out of the picture a good part of the time . Left alone in \textcolor{ent8}{California} , \textcolor{ent8}{Patia} soon filed for divorce . The \textcolor{ent13}{single mother} supported \textcolor{ent13}{her children} with a series of acting jobs in local theater companies ...
\end{tcolorbox}
The blue entities represent specific entities, such as personal names, organizational names, and place names (including full names, abbreviations, nicknames, etc.); The red entity represents a general entity, such as "his wife" or "single mother"; We do not annotate the yellow pronouns, such as "he", "we", "it" and so on. For time entities, we mainly annotate year, month, day, hour, and minute, as shown in the red entity in the following example:
\begin{tcolorbox}
... Personally downhearted and paranoid , Elvis grew bored and dissatisfied . By \textcolor{ent13}{1976} it was almost impossible to get Elvis into a recording studio despite his contractual obligations . Sadly , Elvis Presley died on \textcolor{ent13}{August 16 1977} in the bathroom of a of his Graceland mansion  ...
\end{tcolorbox}

\begin{figure*}[!t]
\centering
\includegraphics[width=0.7\textwidth]{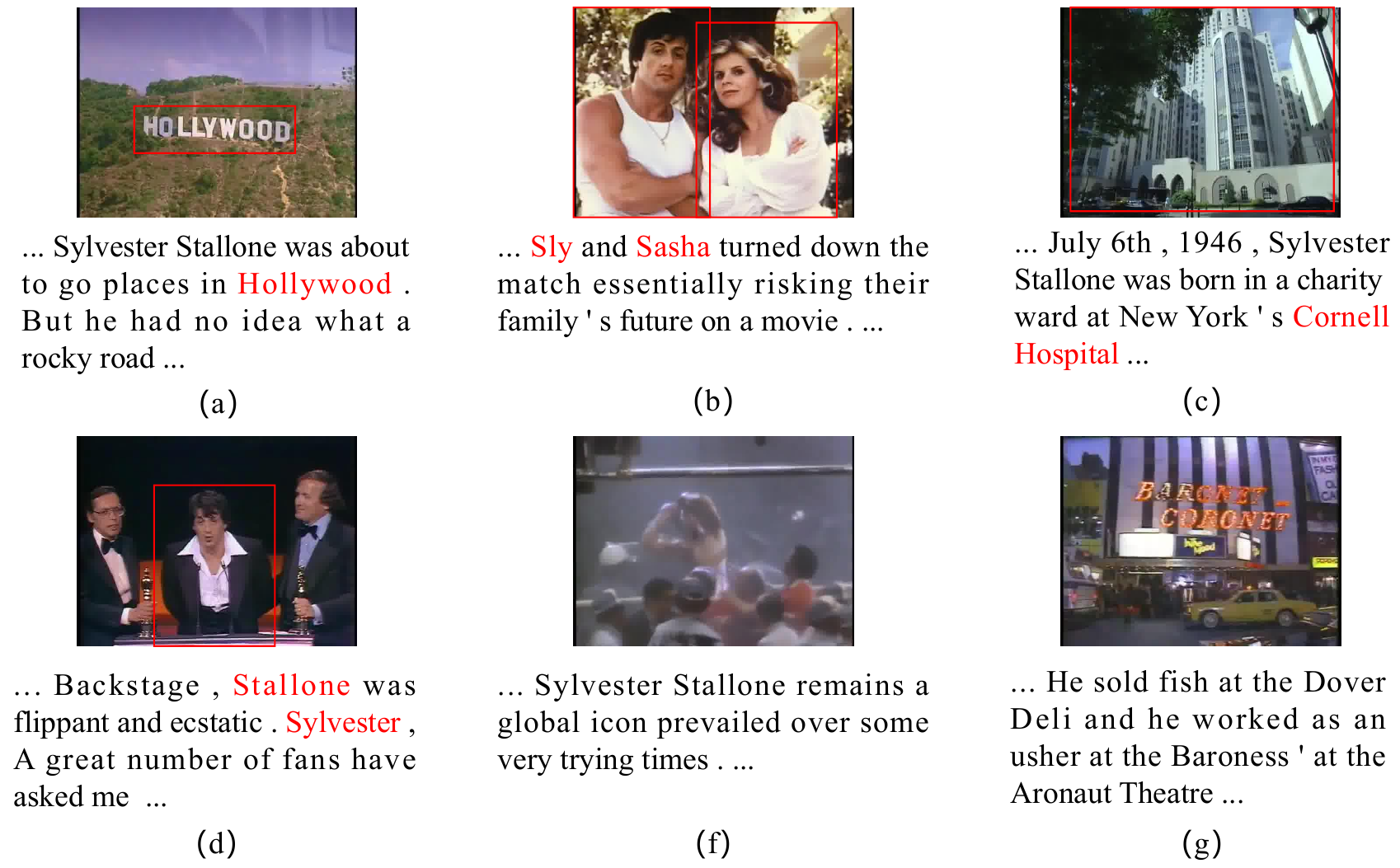}
\caption{Some examples of visual grounding annotations. The red entity in the text corresponds to the visual target in the image.} 
\label{vg_example}
\end{figure*}

\subsection{Entity Chain and Relation Annotation Specification}
Entity chains are equivalent to special entity relation in the annotation process, namely, co-reference relation. 
There are two more points to note during the annotation process: 1) Entity relations are at the entity chain level, so relations between entities from the same entity chain are annotated only once. 2) There is overlap in entity relations, for example, two people are both colleague and couple, then they need to be marked as two relations in the same document, namely PER-PER\_colleague and PER-PER\_couple.
\begin{tcolorbox}
... The chair recognizes \textcolor{ent8}{Glen Campbell} and \textcolor{ent13}{Tanya Tucker} for the singing of our national anthem . Just months after the split with Sarah . \textcolor{ent8}{Glen} ' s affair with 21 year old country singer \textcolor{ent13}{Tanya Tucker} hit the papers and the tabloids The two extended their personal relationship into a professional one recording and touring together ...
\end{tcolorbox}

Entities of the same color constitute an entity chain, where overlapping relations exist among them. Initially, they are colleagues and have a PER-PER\_colleague. Later, they get married, so there is also a PER-PER\_couple.

\subsection{Visual Grounding Annotation Specification}
Visual grounding mainly annotates people, places and organizations. When an entity mentioned in the text appears in the image and the entity is clear enough, we annotate the visual target. Figure \ref{vg_example} shows some examples of what should and should not be annotated. The persons and organizations in images (a)-(c) appear in the text and are very clear, so they need to be annotated. In image (d), although three characters appear, only “Stallone” is mentioned in the text, so only “Stallone” in the middle of the image is annotated. In image (f), “Stallone” appears but is too blurry, so it is not annotated. Similarly, in image (g), the “Baronet coronet” organization is not mentioned in the text, so it is not annotated.

\section{Methodology}
\begin{figure*}[!t]
\centering
\includegraphics[width=0.8\textwidth]{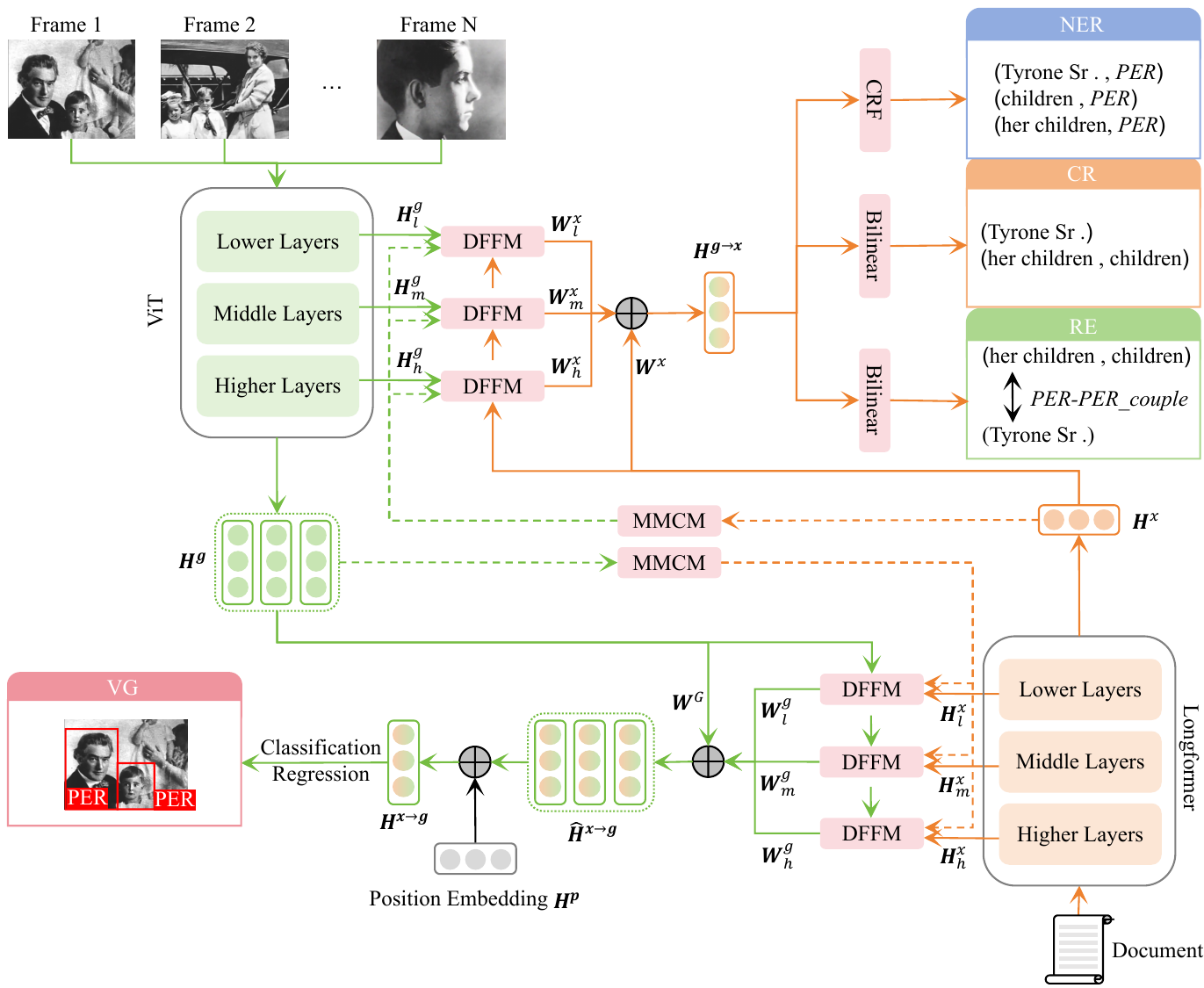}
\caption{The overall architecture of our model. The dashed line indicates execution when the modality is missing. For a detailed introduction to DFFM and MMCM, see Section \ref{DFFM_section} and \ref{MMCM_section}. $\bigoplus$ represents the element-wise summation operation.} 
\label{overall architecture}
\end{figure*}

The overall structure of our model is shown in Figure \ref{overall architecture}, which consists of three parts: encoding layer, denoised feature fusion module and missing modality construction module.

\subsection{Task Definition}
Before introducing each module, we first define our task. Our task goal is to complete four tasks simultaneously, including entity recognition, entity chain extraction, relation extraction and visual grounding. Specifically, given a document $X=\{x_{1},x_{2},\dots,x_{n_{d}}\}\in\mathbb{R}^{n_{d}}$ with a length of $n_{d}$ and a set of images (they come from videos) $G=\{g_{1},g_{2},\dots,g_{n_{g}}\}\in\mathbb{R}^{n_{g}}$ containing $n_{g}$ images, where $x_{i}$ and $g_{i}$ represent the $i$-th word and image respectively.

We are able to identify entities, entity chains, entity relations in text, and target regions in images. These outputs are defined as follows: 

An entity set $E$ containing $n_{e}$ entities,
\begin{equation}
\label{deqn_ex1a}
\begin{gathered}
E=\{e_{1},e_{2},\dots,e_{n_{e}}\}\in\mathbb{R}^{n_{e}}, \\
e_{i} = [s^{e}_{i}, t^{e}_{i}], t^{e}_{i}\in{C}^{ent},\\
\end{gathered}
\end{equation}
where $e_{i}$ represents the $i$-th entity, $s^{e}_{i}$ and $t^{e}_{i}$ represent the span and type of $i$-th entity respectively, ${C}^{ent}$ represents the type set of entities. 

An entity chain set $L$ containing $n_{l}$ entity chains,
\begin{equation}
\label{deqn_ex2a}
\begin{gathered}
L=\{l_{1},l_{2},\dots,l_{n_{l}}\}\in\mathbb{R}^{n_{l}} , \\
l_{i} = \{e_{i,1},e_{i,2},\dots,e_{i,m_{i}}\}\in\mathbb{R}^{m_{i}}, \\
e_{i,j}\in E, 1 \le m_{i} \le n_{e}, \sum_{i=1}^{n_{l}}{m_{i}}=n_{e} \\
\end{gathered}
\end{equation}
where $l_{i}$ represents the $i$-th entity chain, $e_{i, j}$ represents the $j$-th entity in the $i$-th entity chain, and $m_{i}$ represents the length of the $i$-th entity chain. 

A relation set $R$ containing $n_{r}$ pairs of relation triples, 
\begin{equation}
\label{deqn_ex3a}
\begin{gathered}
R=\{r_{1},r_{2},\dots,r_{n_{r}}\}\in\mathbb{R}^{n_{r}}, \\
r_{i} = [l_{i}^{sub}, l_{i}^{obj}, t^{r}_{i}],t^{r}_{i}\in{C}^{rel},\\
l_{i}^{sub}, l_{i}^{obj} \in L, \\
\end{gathered}
\end{equation}
where $r_{i}$ represents the $i$-th relation triple, which contains the subject entity chain $l_{i}^{sub}$, the object entity chain $l_{i}^{obj}$ and the entity relation type $t^{r}_{i}$, ${C}^{rel}$ represents the type set of relations.

A visual region set $V$, The representation of each visual region uses the YOLO \cite{yolo} format, which is a quintuple,
\begin{equation}
\label{deqn_ex4a}
\begin{gathered}
V=\{v_{1},v_{2},\dots,v_{n_{g}}\}\in\mathbb{R}^{n_{g}}, \\
v_{i}=\{v_{i,1},v_{i,2},\dots,v_{i,n_{v}}\}\in\mathbb{R}^{n_{v}}, \\
v_{i,j}=[t^{v}_{i,j},c_{i,j}^{horiz},c_{i,j}^{vert},c_{i,j}^{wd},c_{i,j}^{ht}], t^{v}_{i,j}\in C^{gro},\\
\end{gathered}
\end{equation}
where $v_{i}$ represents the quintuple set of visual targets of the $i$-th image, which contains $n_{v}$ visual targets, $v_{i,j}$ represents the quintuple of the $j$-th visual target of the $i$-th image, $t^{v}_{i,j}$, $c_{i,j}^{horiz}$, $c_{i,j}^{vert}$, $c_{i,j}^{wd}$ and $c_{i,j}^{ht}$ represent the type of visual target, the central horizontal coordinate, central vertical coordinate, width and height of the visual target bounding box respectively, $C^{gro}$ represents the type set of visual areas.

\subsection{Encoding Layer}
Our inputs consist of document-level text $X$ and a set of images $G$, For the text, we use Longformer \cite{longformer} for encoding, which allows us to obtain the text embeddings $\boldsymbol{H}^{x}$,
\begin{equation}
\label{deqn_ex4a}
\begin{gathered}
\boldsymbol{H}^{x}=\{\boldsymbol{h}^{x}_{1},\boldsymbol{h}^{x}_{2},\dots,\boldsymbol{h}^{x}_{n_{x}}\}\in\mathbb{R}^{n_{x}\times d_{h}}, \\
\end{gathered}
\end{equation}
where $\boldsymbol{h}^{x}_{i}\in\mathbb{R}^{d_{h}}$ represents the embedding of the $i$-th word, and $d_{h}$ represents the dimension of the embeddings. We then use ViT \cite{vit} to encode the images. ViT first divides each image into $n_{p}$ patches and then encode the linearized patches, so we can get the image embedding $\boldsymbol{H}^{g}$,
\begin{equation}
\label{deqn_ex4a}
\begin{gathered}
\boldsymbol{H}^{g}=\{\boldsymbol{H}^{g}_{1},\boldsymbol{H}^{g}_{2},\dots,\boldsymbol{H}^{g}_{n_{g}}\}\in\mathbb{R}^{n_{g}\times n_{p}\times d_{h}}, \\
\boldsymbol{H}^{g}_{i}=\{\boldsymbol{h}^{g}_{i,1},\boldsymbol{h}^{g}_{i,2},\dots,\boldsymbol{h}^{g}_{i,n_{p}}\}\in\mathbb{R}^{n_{p}\times d_{h}}, \\
\end{gathered}
\end{equation}
where $\boldsymbol{h}^{g}_{i,j}\in\mathbb{R}^{d_{h}}$ represents the embedding of the $j$-th patch of the $i$-th image.

\subsection{Hierarchical Features}
Some existing studies have shown that different layers of pre-trained models can capture different features. Rogers et al. \cite{bert_layer} found that the lower layers, middle layers, and higher layers of BERT \cite{bert} capture lexical features, syntactic features, and semantic features of text, respectively. Additionally, Dosovitskiy et al. \cite{vit_layer} discovered that the lower layers of ViT capture features such as color, texture, and edges, while the middle layers capture local structures and regional dependencies in images, and the higher layers capture overall object structures, categories, and global dependencies.

Therefore, for a Longformer with $n_{l}$ layers, we can get its low, medium and high features of the text as,
\begin{equation}
\label{deqn_ex4a}
\begin{gathered}
\boldsymbol{H}^{x}_{l}=\frac{3}{n_{l}}\sum_{i=1}^{n_{l}/3}(\boldsymbol{H}^{x}_{i}),\\
\boldsymbol{H}^{x}_{m}=\frac{3}{n_{l}}\sum_{i=n_{l}/3}^{2*n_{l}/3}(\boldsymbol{H}^{x}_{i}),\\
\boldsymbol{H}^{x}_{h}=\frac{3}{n_{l}}\sum_{i=2*n_{l}/3}^{n_{l}}(\boldsymbol{H}^{x}_{i}),\\
\end{gathered}
\end{equation}
where $\boldsymbol{H}^{x}_{i}\in\mathbb{R}^{n_{x}\times d_{h}}$ represents the feature of the $i$-th layer. Similarly, we can get the low, medium and high features of the image as $\boldsymbol{H}^{g}_{l}$, $\boldsymbol{H}^{g}_{m}$ and $\boldsymbol{H}^{g}_{h}$.

\begin{figure*}[!t]
\centering
\includegraphics[width=0.7\textwidth]{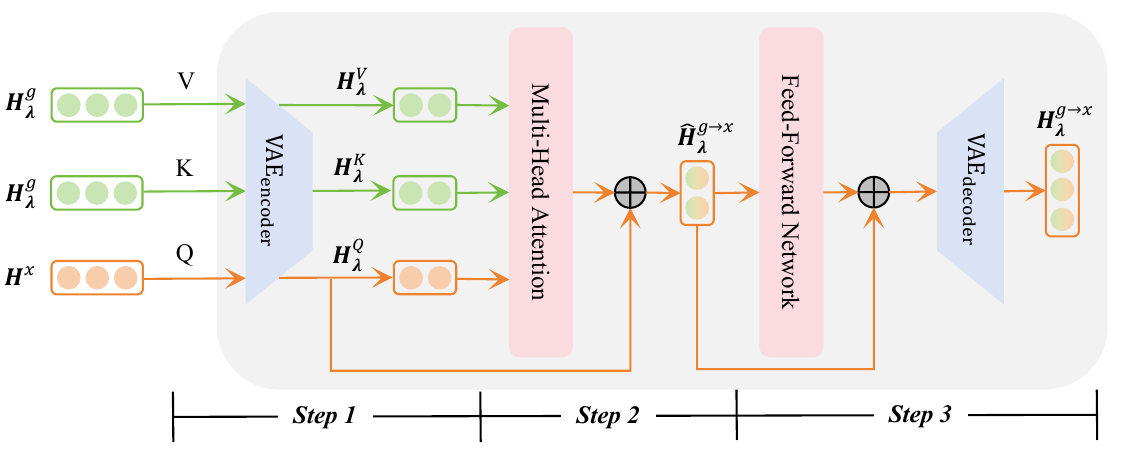}
\caption{The architecture diagram of our denoised feature fusion module (DFFM). We take image integration into text as an example. \textbf{Step 1:} $\boldsymbol{H}^{x}$ (as Q) and $\boldsymbol{H}^{g}_{\lambda}$ (as K and V) are reduced in dimension through the VAE encoder to obtain $\boldsymbol{H}^{Q}_{\lambda}$, $\boldsymbol{H}^{K}_{\lambda}$ and $\boldsymbol{H}^{V}_{\lambda}$. \textbf{Step 2:} $\boldsymbol{H}^{Q}_{\lambda}$, $\boldsymbol{H}^{K}_{\lambda}$ and $\boldsymbol{H}^{V}_{\lambda}$ are processed through the multi-head attention mechanism and residual network to obtain $\boldsymbol{\hat{H}}^{g\to x}_{\lambda}$. \textbf{Step 3:} After $\boldsymbol{\hat{H}}^{g\to x}_{\lambda}$ passes through FFN and residual network, the dimension is increased through the decoder of VAE to obtain the final fusion feature $\boldsymbol{H}^{g\to x}_{\lambda}$.} 
\label{DFFM}
\end{figure*}

\subsection{Denoised Feature Fusion Module (DFFM)}
\label{DFFM_section}

After obtaining the hierarchical features, we fuse them into other modalities through the denoised feature fusion module (DFFM). The structure of DFFM is shown in Figure \ref{DFFM}, which contains a variational auto-encoder (VAE) \cite{VAE}, a multi-head attention layer and a feed-forward network (FFN).  
Considering the presence of certain noise in both images and entities, in order to reduce the impact of noise during feature fusion, we use VAE to map features to latent space for fusion \cite{vae_latent}. We believe that redundant noise information has been removed from the features in latent space.

\textbf{Image Integrated into Text:} We use the text feature $\boldsymbol{H}^{x}$ as the query (Q) of the multi-head attention mechanism, and the low, medium and high features of the image $\boldsymbol{H}^{g}_{\lambda}\in\mathbb{R}^{n_{g}\times n_{p}\times d_{h}}$ as the key (K) and value (V), where $\lambda\in\{l,m,h\}$. First, Q, K and V are reduced in dimension through the VAE encoder,
\begin{equation}
\label{deqn_endocer}
\begin{gathered}
\boldsymbol{H}^{Q}_{\lambda}=\text{Encoder}_{\lambda}(\boldsymbol{H}^{x}),\\
\boldsymbol{H}^{K}_{\lambda}=\text{Encoder}_{\lambda}(\boldsymbol{W}^{K}_{\lambda}\boldsymbol{H}^{g}_{\lambda}),\\
\boldsymbol{H}^{V}_{\lambda}=\text{Encoder}_{\lambda}(\boldsymbol{W}^{V}_{\lambda}\boldsymbol{H}^{g}_{\lambda}),\\
\end{gathered}
\end{equation}
where $\boldsymbol{H}^{Q}_{\lambda}\in\mathbb{R}^{n_{x}\times d_{vae}}$ and $\boldsymbol{H}^{K}_{\lambda}$, $\boldsymbol{H}^{V}_{\lambda}\in\mathbb{R}^{n_{g}\times n_{p}\times d_{vae}}$, $d_{vae}$ represents the dimension after VAE dimensionality reduction, and $\boldsymbol{W}^{K}_{\lambda}$, $\boldsymbol{W}^{V}_{\lambda}\in\mathbb{R}^{d_{h}\times d_{h}}$ represent the trainable weights. Then input it into the multi-head attention mechanism \cite{attention} and residual network \cite{residual} to get:
\begin{equation}
\label{deqn_ex4a}
\begin{gathered}
\boldsymbol{\hat{H}}^{g\to x}_{\lambda}=\boldsymbol{H}^{Q}_{\lambda}+\text{MultiHeadAttention}_{\lambda}(\boldsymbol{H}^{Q}_{\lambda},\boldsymbol{H}^{K}_{\lambda},\boldsymbol{H}^{V}_{\lambda}),\\
\end{gathered}
\end{equation}
where $\boldsymbol{\hat{H}}^{g\to x}_{\lambda}\in\mathbb{R}^{n_{x}\times d_{ave}}$, and $g\to x$ means that image features are integrated into text features. Then $\boldsymbol{\hat{H}}^{g\to x}_{\lambda}$ is input into FFN and residual network, and the dimension is increased through the VAE decoder to obtain the fusion feature,
\begin{equation}
\label{deqn_decoder}
\begin{gathered}
\boldsymbol{H}^{g\to x}_{\lambda}=\text{Decoder}_{\lambda}(\boldsymbol{\hat{H}}^{g\to x}_{\lambda}+\text{FFN}_{\lambda}(\boldsymbol{\hat{H}}^{g\to x}_{\lambda})),\\
\end{gathered}
\end{equation}
where $\boldsymbol{H}^{g\to x}_{\lambda}\in\mathbb{R}^{n_{x}\times d_{h}}$. Finally, the low, medium and high-level fusion features are weighted to obtain the hierarchical text features,
\begin{equation}
\label{deqn_hierarchical}
\begin{gathered}
\boldsymbol{H}^{g\to x}=\sum_{\lambda}\boldsymbol{W}^{x}_{\lambda}\boldsymbol{H}^{g\to x}_{\lambda}+\boldsymbol{W}^{x}\boldsymbol{H}^{x},\\
\end{gathered}
\end{equation}
where $\boldsymbol{W}^{x},\boldsymbol{W}^{x}_{\lambda}\in\mathbb{R}^{d_{h}\times d_{h}}$ represents the trainable weights. 

\textbf{Text Integrated into Images:} We use the image feature $\boldsymbol{H}^{g}$ as the Q, and the low, medium and high features of the text $\boldsymbol{H}^{x}_{\lambda}\in\mathbb{R}^{n_{x}\times d_{h}}$ as the K and V. Then, according to Formula \ref{deqn_endocer} to \ref{deqn_hierarchical}, we get the hierarchical image feature $\boldsymbol{\hat{H}}^{x\to g}\in\mathbb{R}^{n_{g}\times n_{p}\times d_{h}}$ containing intra-frame information. Then we fuse the intra-frame features using average pooling ($\text{AvgPool}$). In addition, we introduce positional embeddings $\boldsymbol{H}^{p}$ to account for the inter-image frame temporal information. Fianlly, we obtain the hierarchical image features $\boldsymbol{H}^{x\to g}\in\mathbb{R}^{n_{g}\times d_{h}}$,
\begin{equation}
\label{deqn}
\begin{gathered}
\boldsymbol{H}^{x\to g}=\text{AvgPool}(\boldsymbol{\hat{H}}^{x\to g})+\boldsymbol{H}^{p}.\\
\end{gathered}
\end{equation}

\begin{figure}[!t]
\centering
\includegraphics[width=0.4\textwidth]{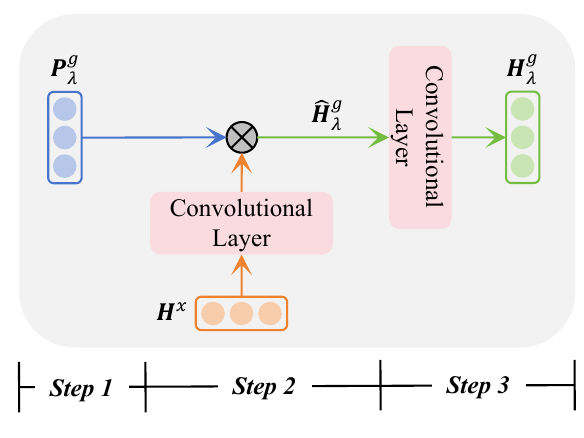}
\caption{The architecture diagram of our missing modality construction module (MMCM). We take the generation of the missing image modality from the text modality as an example. \textbf{Step 1:} Randomly initialize the missing image modality prompt $\boldsymbol{P}_{\lambda}^{g}$. \textbf{Step 2:} The text embeddings $\boldsymbol{H}^{x}$ are passed through a convolutional layer and concatenated with $\boldsymbol{P}_{\lambda}^{g}$ to obtain $\boldsymbol{\hat{H}}_{\lambda}^{g}$, where 
$\bigotimes$ denotes the concatenation operation.
\textbf{Step 3:} After $\boldsymbol{\hat{H}}_{\lambda}^{g}$ passes through the convolutional layer, the missing image modality feature $\boldsymbol{H}_{\lambda}^{g}$ is obtained.} 
\label{MMCM}
\end{figure}

\subsection{Missing Modality Construction Module (MMCM)}
\label{MMCM_section}
In non-ideal situations, modal information is incomplete. When a modal information is missing, our DFFM will lose its function, resulting in a decrease in the overall performance of the model. Therefore, we designed a missing modality construction module (MMCM). The structure of MMCM is shown in Figure \ref{MMCM}. 
We were inspired by Guo et al. \cite{missing_prompt_msa} to construct a true modal generation module based on convolutional layers. The reason for using convolutional layers is due to their locality and translational invariance, which means they can extract local features of modalities without being affected by position.

\textbf{Text to Image:} First, we randomly initialize a missing image modality prompt $\boldsymbol{P}^{g}_{\lambda}\in\mathbb{R}^{l_{p}\times d_{h}}$, where $l_{p}$ represents the length of the prompt. Then we concatenate the text embedding $\boldsymbol{H}^{x}$ with $\boldsymbol{P}^{g}_{\lambda}$ after passing it through the convolutional layer (Conv) to obtain $\boldsymbol{\hat{H}}^{g}_{\lambda}$, 
\begin{equation}
\label{deqn}
\begin{gathered}
\boldsymbol{\hat{H}}^{g}_{\lambda}=[\boldsymbol{P}^{g}_{\lambda};\text{ReLU}(\text{Conv}(\boldsymbol{H}^{x}))],\\
\end{gathered}
\end{equation}
where $\boldsymbol{\hat{H}}^{g}_{\lambda}\in\mathbb{R}^{(l_{p}+n_{x})\times d_{h}}$, $[\cdot;\cdot]$ represents the concatenation operation. Finally, $\boldsymbol{\hat{H}}^{g}_{\lambda}$ passes through the convolutional layer to obtain the missing image modality feature $\boldsymbol{H}^{g}_{\lambda}$,
\begin{equation}
\label{deqn}
\begin{gathered}
\boldsymbol{H}^{g}_{\lambda}=\text{ReLU}(\text{Conv}_{\lambda}(\boldsymbol{\hat{H}}^{g}_{\lambda}))].\\
\end{gathered}
\end{equation}

\textbf{Image to Text:} Similarly, we can generate missing text modality features $\boldsymbol{H}^{x}_{\lambda}$ through image embedding $\boldsymbol{H}^{g}$ and randomly initialized missing text modality prompt $\boldsymbol{P}_{\lambda}^{x}$.

In this way, we can input the non-missing modal features and the generated missing modal features into DFFM (see Section \ref{DFFM_section} for details) to obtain the hierarchical features.

\subsection{Output}
For the NER task, we regard the task as a classification task and use BIO tags to identify entities. Therefore, we use CRF \cite{crf} to classify the hierarchical text feature $\boldsymbol{H}^{g\to x}$ and obtain the prediction result $\boldsymbol{y}^{ent}$,
\begin{equation}
\label{deqn}
\begin{gathered}
\boldsymbol{y}^{ent}=\text{CRF}(\boldsymbol{W}^{ent}\boldsymbol{H}^{g\to x}),\\
\end{gathered}
\end{equation}
where $\boldsymbol{W}^{ent}\in\mathbb{R}^{d_{h}\times l^{ent}}$ represents the trainable, $l^{ent}$ represents the length of the BIO tag set. 

For the CR task, we regard the task as a binary classification task and predict the entity chain by predicting whether there is a co-reference relation between entity pairs. We get the representation ($\boldsymbol{H}_{sub}^{ent},\boldsymbol{H}_{obj}^{ent}$) of the entity pair from $\boldsymbol{H}^{g\to x}$ and pass it through the Blinear layer to get the prediction result $\boldsymbol{y}^{cha}$,
\begin{equation}
\label{deqn}
\begin{gathered}
\boldsymbol{y}^{cha}=\text{argmax}(\boldsymbol{W}^{cha}(\boldsymbol{H}_{sub}^{ent}\circledast \boldsymbol{H}_{obj}^{ent})),\\
\end{gathered}
\end{equation}
where $\boldsymbol{W}^{cha}\in\mathbb{R}^{d_{h}\times 2}$ represents the trainable, $\circledast$ represents Kronecker product. 

For the RE task, we regard the task as a multi-classification task. We get the representation ($\boldsymbol{H}_{sub}^{cha},\boldsymbol{H}_{obj}^{cha}$) of the entity chain pair from $\boldsymbol{H}^{g\to x}$ and pass it through the Blinear layer to get the prediction result $\boldsymbol{y}^{rel}$,
\begin{equation}
\label{deqn}
\begin{gathered}
\boldsymbol{y}^{rel}=\text{argmax}(\boldsymbol{W}^{rel}(\boldsymbol{H}_{sub}^{cha}\circledast \boldsymbol{H}_{obj}^{cha})),\\
\end{gathered}
\end{equation}
where $\boldsymbol{W}^{rel}\in\mathbb{R}^{d_{h}\times l_{rel}}$ represents the trainable, $l^{rel}$ represents the length of the relation type set. 

For the VG task, we regard the recognition of the types of visual regions as a multi-class classification task, while the recognition of the boundaries of visual regions is treated as a regression task. We first map the hierarchical image features $\boldsymbol{H}^{x\to g}$ through linear layers before proceeding with classification and regression to obtain the prediction results $\boldsymbol{y}^{gro_{t}}$ and $\boldsymbol{y}^{gro_{b}}$,
\begin{equation}
\label{deqn}
\begin{gathered}
\boldsymbol{y}^{gro_{t}}=\text{argmax}(\boldsymbol{W}^{gro_{t}}\boldsymbol{H}^{x\to g}),\\
\boldsymbol{y}^{gro_{b}}=\sigma(\boldsymbol{W}^{gro_{b}}\boldsymbol{H}^{x\to g}),\\
\end{gathered}
\end{equation}
where $\boldsymbol{W}^{gro_{t}}\in\mathbb{R}^{d_{h}\times l_{gro}}$ and $\boldsymbol{W}^{gro_{b}}\in\mathbb{R}^{d_{h}\times 4}$ represent the trainable, $l^{gro}$ represents the length of the visual area type set, $\sigma$ represents the Sigmoid activation function. 

\subsection{Learning}
We define the golden result of each task as $\boldsymbol{\hat{y}}^{\gamma}$, where $\gamma\in\{ent,cha,rel,gro_{t},gro_{b}\}$. For the classification task, our training goal is to minimize the negative log-likelihood loss between the predicted probability of sequence and the probability of the corresponding gold sequence. For the NER task, the loss is:
\begin{equation}
\label{deqn_ex11a}
\mathcal{L}^{ent}=-\frac{1}{n_{x}}\sum_{i=1}^{n_{x}}\sum_{j=1}^{l^{ent}}\boldsymbol{\hat{y}}^{ent}_{i,j}\text{log}\boldsymbol{y}_{i,j}^{ent} \,.
\end{equation}
For the CR task, the loss is:
\begin{equation}
\label{deqn_ex11a}
\mathcal{L}^{cha}=-\frac{1}{n_{ep}}\sum_{i=1}^{n_{ep}}\sum_{j=1}^{2}\boldsymbol{\hat{y}}^{cha}_{i,j}\text{log}\boldsymbol{y}_{i,j}^{cha} \,,
\end{equation}
where $n_{ep}$ represents the number of entity pairs. For the RE task, the loss is:
\begin{equation}
\label{deqn_ex11a}
\mathcal{L}^{rel}=-\frac{1}{n_{cp}}\sum_{i=1}^{n_{cp}}\sum_{j=1}^{l^{rel}}\boldsymbol{\hat{y}}^{rel}_{i,j}\text{log}\boldsymbol{y}_{i,j}^{rel} \,,
\end{equation}
where $n_{cp}$ represents the number of entity chain pairs. The visual area type prediction loss for the VG task is:
\begin{equation}
\label{deqn_ex11a}
\mathcal{L}^{gro_{t}}=-\frac{1}{n_{va}}\sum_{i=1}^{n_{va}}\sum_{j=1}^{l^{gro}}\boldsymbol{\hat{y}}^{gro_{t}}_{i,j}\text{log}\boldsymbol{y}_{i,j}^{gro_{t}} \,,
\end{equation}
where $n_{va}$ represents the number of visual areas.

For the regression task, our goal is to minimize the minimum distance between the predicted result and the corresponding golden result. Here, we use the mean absolute error (MAE), and we can get the visual region boundary prediction loss for the VG task:
\begin{equation}
\label{deqn_ex11a}
\mathcal{L}^{gro_{b}}=\frac{1}{n_{va}}\sum_{i=1}^{n_{va}}\lvert \boldsymbol{\hat{y}}^{gro_{b}}_{i}-\boldsymbol{y}_{i}^{gro_{b}} \rvert .
\end{equation}

Finally we can get the total loss as:
\begin{equation}
\label{deqn_ex11a}
\mathcal{L}=\sum_{\gamma}\alpha^{\gamma}\mathcal{L}^{\gamma},
\end{equation}
where $\alpha^{\gamma}$ represents the hyper-parameter set.

\section{Experiment Setups}
\subsection{Baselines}
\textbf{One-Stream architecture (OS)} \cite{OS} is designed to process video data through a single channel.
\textbf{Two-Stream architecture (TS)} \cite{TS} processes video data through two independent channels, usually one channel is used to process spatial information (such as static frames) and the other channel is used to process temporal information (such as optical flow or inter-frame motion).
\textbf{ViViT} \cite{vivit} is a Transformer model specifically designed for video understanding. It treats the video as a three-dimensional data cube (time, width, height) and processes it using the Transformer architecture.
\textbf{VideoMAE} \cite{videomae} is a self-supervised learning model for video understanding. Its design is inspired by the image MAE (Masked Autoencoder), which aims to learn the potential representation of video data by masking some frames in the video sequence.
\textbf{MDocRE-HN} \cite{re_video_data} is a hierarchical framework to learn interactions between different dependency levels and a text-guided transformer architecture that incorporates both text and video modalities.
\textbf{Video-LLaMA-2} \cite{video-llama} is a multimodal large language model (MLLM) that aims to combine video input with text generation to enhance the user-model interaction experience. The model is based on the LLaMA architecture and integrates the advantages of visual transformers and text transformers to achieve understanding and generation of video content.
\textbf{Video-ChatGPT} \cite{videochatgpt} is a MLLM that merges a video-adapted visual encoder with an LLM. The resulting model is capable of understanding and generating detailed conversations about videos.

\begin{table*}[!t]
\fontsize{7}{9}\selectfont
\setlength{\tabcolsep}{0.8mm}
\centering
\caption{Comparison of our model with the baseline model. We only report the F1 score and variance (values in brackets) for each task. \textcolor{bc}{\textbf{green score}} indicates the best result in each column. \textcolor{uc}{blue score} indicates the second best result in each column. (Avg.:Average performance of the four tasks)}
\resizebox{0.8\textwidth}{!}{
\begin{tabular}{l l c c c c c c c c c c}
\hline
&&\phantom{}&\bf Ent.&\phantom{}&\bf Cha.&\phantom{}&\bf Rel.&\phantom{}&\bf Gro.&\phantom{}&\bf Avg.\\
\hline
\multirow{8}{*}{\bf EN}&VideoMAE+Longformer+Os&\phantom{}&89.65\mysize{(0.01)}&\phantom{}&57.43\mysize{(0.03)}&\phantom{}&27.72\mysize{(0.01)}&\phantom{}&27.23\mysize{(0.01)}&\phantom{}&50.51\\
&VideoMAE+Longformer+Ts&\phantom{}&90.55\mysize{(0.01)}&\phantom{}&58.20\mysize{(0.01)}&\phantom{}&28.94\mysize{(0.01)}&\phantom{}&29.54\mysize{(0.01)}&\phantom{}&51.81\\
&ViViT+Longformer+Os&\phantom{}&91.11\mysize{(0.01)}&\phantom{}&58.28\mysize{(0.01)}&\phantom{}&27.47\mysize{(0.14)}&\phantom{}&27.09\mysize{(0.22)}&\phantom{}&50.99\\
&ViViT+Longformer+Ts&\phantom{}&\textcolor{bc}{\textbf{92.03}}\mysize{(0.01)}&\phantom{}&\textcolor{uc}{59.23}\mysize{(0.01)}&\phantom{}&28.51\mysize{(0.40)}&\phantom{}&30.87\mysize{(0.30)}&\phantom{}&52.66\\
&MDocRE-HN&\phantom{}&91.20\mysize{(0.01)}&\phantom{}&58.09\mysize{(0.27)}&\phantom{}&\textcolor{uc}{29.15}\mysize{(0.23)}&\phantom{}&\textcolor{uc}{32.15}\mysize{(0.69)}&\phantom{}&\textcolor{uc}{52.65}\\
\cdashline{2-12}
&Video-LLaMA-2&\phantom{}&13.04&\phantom{}&26.51&\phantom{}&0.29&\phantom{}&8.63&\phantom{}&12.12\\
&Video-ChatGPT&\phantom{}&35.69&\phantom{}&36.21&\phantom{}&3.23&\phantom{}&14.75&\phantom{}&22.47\\
\cdashline{2-12}
&Our&\phantom{}&\textcolor{uc}{91.24}\mysize{(0.01)}&\phantom{}&\textcolor{bc}{\textbf{59.67}}\mysize{(0.01)}&\phantom{}&\textcolor{bc}{\textbf{29.78}}\mysize{(0.02)}&\phantom{}&\textcolor{bc}{\textbf{34.52}}\mysize{(0.03)}&\phantom{}&\textcolor{bc}{\textbf{53.80}}\\
\hline
\multirow{8}{*}{\bf ZH}&VideoMAE+Longformer+Os&\phantom{}&83.15\mysize{(0.51)}&\phantom{}&60.72\mysize{(0.64)}&\phantom{}&26.96\mysize{(0.36)}&\phantom{}&20.28\mysize{(0.02)}&\phantom{}&47.78\\
&VideoMAE+Longformer+Ts&\phantom{}&87.96\mysize{(0.01)}&\phantom{}&64.03\mysize{(0.01)}&\phantom{}&29.82\mysize{(0.05)}&\phantom{}&22.13\mysize{(0.06)}&\phantom{}&50.99\\
&ViViT+Longformer+Os&\phantom{}&87.66\mysize{(0.01)}&\phantom{}&64.34\mysize{(0.01)}&\phantom{}&30.67\mysize{(0.01)}&\phantom{}&21.02\mysize{(0.04)}&\phantom{}&50.92\\
&ViViT+Longformer+Ts&\phantom{}&\textcolor{bc}{\textbf{88.00}}\mysize{(0.01)}&\phantom{}&65.55\mysize{(0.01)}&\phantom{}&32.21\mysize{(0.10)}&\phantom{}&\textcolor{bc}{\textbf{22.99}}\mysize{(0.25)}&\phantom{}&52.19\\
&MDocRE-HN&\phantom{}&87.85\mysize{(0.01)}&\phantom{}&\textcolor{uc}{66.04}\mysize{(0.67)}&\phantom{}&\textcolor{uc}{33.97}\mysize{(0.48)}&\phantom{}&21.82\mysize{(0.09)}&\phantom{}&\textcolor{uc}{52.42}\\
\cdashline{2-12}
&Video-LLaMA-2&\phantom{}&21.05&\phantom{}&27.33&\phantom{}&0.19&\phantom{}&1.93&\phantom{}&12.63\\
&Video-ChatGPT&\phantom{}&40.99&\phantom{}&30.35&\phantom{}&1.50&\phantom{}&5.43&\phantom{}&19.56\\
\cdashline{2-12}
&Our&\phantom{}&\textcolor{uc}{87.98}\mysize{(0.01)}&\phantom{}&\textcolor{bc}{\textbf{67.80}}\mysize{(0.04)}&\phantom{}&\textcolor{bc}{\textbf{36.89}}\mysize{(0.01)}&\phantom{}&\textcolor{uc}{22.42}\mysize{(0.02)}&\phantom{}&\textcolor{bc}{\textbf{53.77}}\\
\hline
\end{tabular}
}
\label{tab:main_result}
\end{table*}

\subsection{Evaluation Metrics}
For the four tasks of entity recognition, entity chain extraction, relation extraction and visual grounding, we all use precision (P), recall (R) and F1.

In the entity recognition task, if the token sequence and type of a predicted entity are exactly the same as those of a
golden entity, the predicted entity is regarded as true-positive.

In the entity chain extraction task, we follow Tang et al. \cite{chain_Metrics} to use the average of three indicators: MUC, B$^3$ and CEAF to evaluate the prediction entity chain results. 

In the relation extraction task, we evaluate entity relations at the chain level. When given a predicted relation triplet $r_{pred}$ and a golden relation triplet $r_{gold}$, then when $l_{pred}^{sub} \cap l_{gold}^{sub} \neq \emptyset$, $l_{pred}^{obj} \cap l_{gold}^{obj} \neq \emptyset$ and $t_{pred}^{r} = t_{gold}^{r}$, the predicted relation triplet is regarded as true-positive.

In the visual grounding task, we follow the evaluation method of Yu et al. \cite{gmner}. When the IoU score of the predicted visual area and the golden visual area is greater than 0.5 and the type is the same, then the predicted visual area is regarded as true-positive.

\subsection{Implementation Details}
In the main experiments, we randomly combine video pre-trained models (such as ViViT and VideoMAE), text pre-trained models (such as Longformer), and modality fusion methods (such as OS and TS) as baseline models. 
For entity chain extraction and relation extraction tasks, their entity (chain) pairs are both derived from golden entity (chain) pairs.

In all subsequent experiments, our model uses the text pre-training models \texttt{longformer-base-4096} and \texttt{xlm-roberta-longformer-base-4096} for English and Chinese data respectively, and the visual pre-training model used by our model is \texttt{vit-base-patch16-224}.

Our main experiments will divide the train/dev/test sets into three equal parts, which are set to full modality, missing text modality, and missing visual modality. The LLMs are all tested under the zero-shot setting. Our experimental results are the average of three results. 
The training epochs of our model are 20, the batch size is 1, the learning rate of the pre-trained model is 5e-6, and the learning rate of other modules is 1e-3.
Our model is implemented with PyTorch and trained with a NVIDIA RTX 3090 GPU.


\section{Results and Analysis}
\subsection{Main Comparisons}
The comparison between our model and the baseline model is shown in Table \ref{tab:main_result}. As can be seen from the table, 1) our model has the best effect on average results compared with the baseline model, and it is 1.15\% (53.80\%-52.65\%) and 1.35\% (53.77\%-52.42\%) higher than MDocRE-HN on English and Chinese datasets respectively; 2) the performance of the large language model on our dataset is very poor, both in terms of individual tasks and average results, it is far lower than our model; 3) in the entity recognition task, our model is lower than ViViT+Longformer+Ts, which may be because the entity recognition task does not require deeper reasoning than the entity chain extraction and relation extraction tasks, and the hierarchical modal fusion does not bring obvious performance improvement.

\begin{table*}[!t]
\fontsize{7}{9}\selectfont
\setlength{\tabcolsep}{0.8mm}
\centering
\caption{Ablation experiment.}
\resizebox{0.7\textwidth}{!}{
\begin{tabular}{l l c c c c c c c c c c}
\hline
&&\phantom{}&\bf Ent.&\phantom{}&\bf Cha.&\phantom{}&\bf Rel.&\phantom{}&\bf Gro.&\phantom{}&\bf Avg.\\
\hline
\multirow{3}{*}{\bf EN}
&Best Setting&\phantom{}&\textcolor{bc}{\textbf{91.24}}\mysize{(0.01)}&\phantom{}&\textcolor{bc}{\textbf{59.67}}\mysize{(0.01)}&\phantom{}&\textcolor{bc}{\textbf{29.78}}\mysize{(0.02)}&\phantom{}&\textcolor{bc}{\textbf{34.52}}\mysize{(0.03)}&\phantom{}&\textcolor{bc}{\textbf{53.80}}\\
\cdashline{2-12}
&\quad $w/o$ MMCM &\phantom{} & 90.03\mysize{(0.01)}&\phantom{}& 	{59.45}\mysize{(0.01)}&\phantom{} 	&28.17\mysize{(0.01)} 	&\phantom{}&34.45\mysize{(0.01)}&\phantom{}& 	53.03\\ 
&\quad $w/o$ DFFM &\phantom{}&{90.68}\mysize{(0.01)}&\phantom{}&	58.74\mysize{(0.02)}&\phantom{}& 	29.00\mysize{(0.01)} &\phantom{}&	32.68\mysize{(0.01)}&\phantom{}& 	52.78\\ 
\hline
\multirow{3}{*}{\bf ZH}&Best Setting&\phantom{}&\textcolor{bc}{\textbf{87.98}}\mysize{(0.01)}&\phantom{}&\textcolor{bc}{\textbf{67.80}}\mysize{(0.04)}&\phantom{}&\textcolor{bc}{\textbf{36.89}}\mysize{(0.01)}&\phantom{}&\textcolor{bc}{\textbf{22.42}}\mysize{(0.02)}&\phantom{}&\textcolor{bc}{\textbf{53.77}}\\
\cdashline{2-12}
&\quad $w/o$ MMCM &\phantom{} & 87.08\mysize{(0.18)}&\phantom{}& 	67.44\mysize{(0.34)}&\phantom{} 	&35.88\mysize{(2.32)} 	&\phantom{}&22.02\mysize{(6.12)}&\phantom{}& 	53.11\\ 
&\quad $w/o$ DFFM &\phantom{}&87.77\mysize{(0.01)}&\phantom{}&	65.90\mysize{(0.01)}&\phantom{}& 31.80\mysize{(0.01)} &\phantom{}&21.99\mysize{(0.01)}&\phantom{}& 	51.86\\ 
\hline
\end{tabular}
}
\label{tab:Ablation}
\end{table*}
\subsection{Ablation Analysis}
We conducted an ablation experiment, and the results are shown in Table \ref{tab:Ablation}. First, we removed MMCM, and for the missing modalities, we used blank filling. In English and Chinese, the performance of our model decreased in average results, down 0.77\% (53.80\%-53.03\%) and 0.66\% (53.77\%-53.11\%) respectively. This shows that our proposed MMCM can alleviate the problems caused by missing modalities to a certain extent. Then we removed DFFM, and the performance also decreased, down 1.02\% (53.80\%-52.78\%) and 1.91\% (53.77\%-51.86\%) respectively in average results. This is because after removing DFFM, we no longer perform feature fusion between modalities, and our model degenerates into a singlemodal model. Due to the lack of hierarchical modal features, the model performance decreases. This is also the reason why the performance degradation caused by removing DFFM is greater than that caused by removing MMCM.

\begin{figure*}[!t]
\centering
\includegraphics[width=1\textwidth]{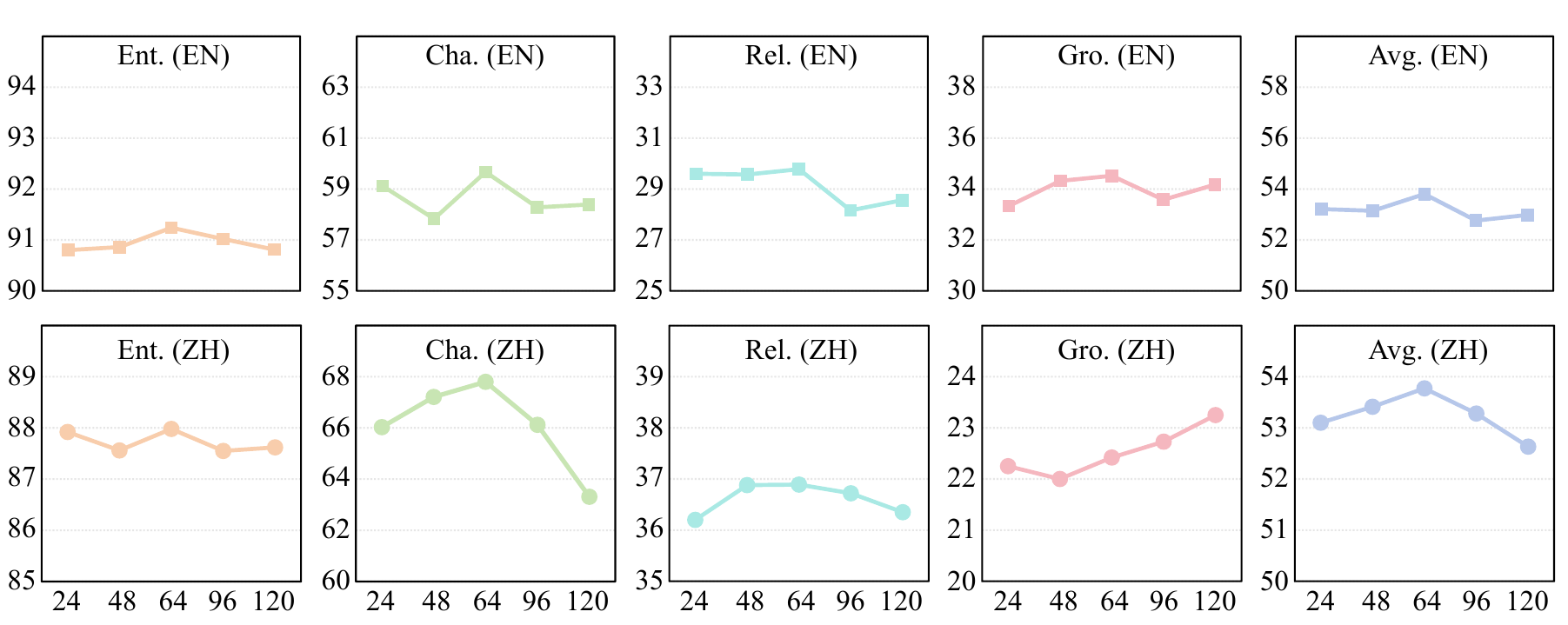}
\caption{The effect of prompt length on the model. The horizontal axis and vertical axis represent the prompt length $n_{p}$ and the F1 value of the model respectively.} 
\label{prompt_len}
\end{figure*}

\subsection{Effect of Prompt Length on the Model}
We analyzed the impact of prompt length on model performance. Detailed results are shown in Figure \ref{prompt_len}. From the average results, when the prompt length is longer or shorter, the model performance decreases. This may be because when the prompt is longer, too many parameters will lead to overfitting; when the prompt is shorter, it may not contain enough information. From the results of individual tasks, the change in prompt length has little impact on the entity recognition task. In addition, we found that the model performance in the visual grounding task under Chinese data improves as the prompt length increases. This may be due to our training method. Our model tries to minimize the overall loss on the four tasks during training. When the prompt increases, the entity chain extraction task has a significant decrease, which leads to improved performance in the visual grounding task.

\subsection{Effect of Missing Modality Ratio on the Model}
We also analyzed the impact of changes in the missing modality ratio on the model. The missing modality ratio here refers to the ratio of the total number of samples of the missing text modality and the missing image modality to the total number of samples, and the number of missing text modality samples and missing image modality samples is the same.

\begin{figure*}[!t]
\centering
\includegraphics[width=1\textwidth]{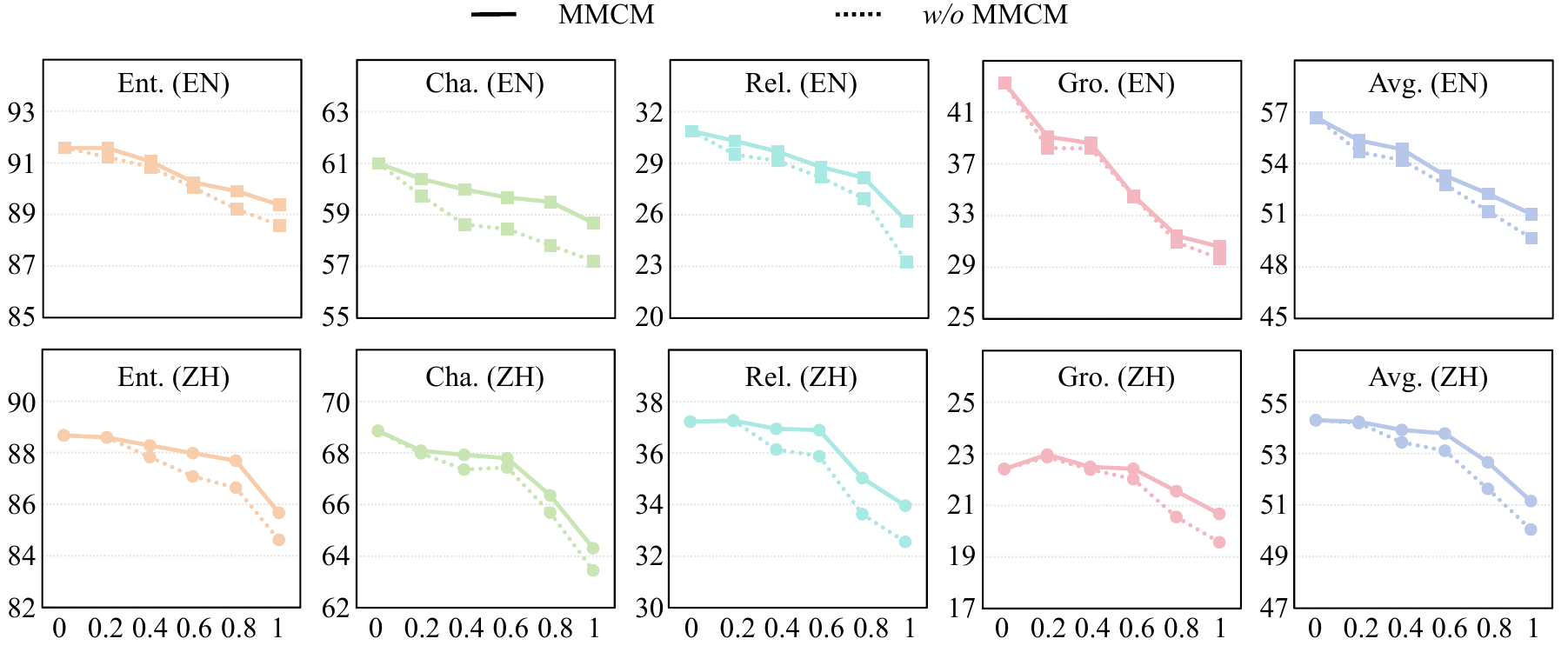}
\caption{The effect of missing modality ratio on the model. The horizontal axis and vertical axis represent the missing modality ratio and the F1 value of the model, respectively.} 
\label{miss_rate}
\end{figure*}

Detailed results are shown in Figure \ref{miss_rate}. From the average results, the performance of our model in both languages decreases with the increase of the missing ratio, and the same is true for individual tasks. Although our results will decrease with the increase of the missing ratio, the downward trend is slower than when MMCM is removed. This also shows that our MMCM can alleviate the performance degradation caused by the missing modality to a certain extent.

\begin{table*}[!t]
\fontsize{8}{10}\selectfont
\setlength{\tabcolsep}{0.8mm}
\centering
\caption{Comparison of predictions for two data samples. Entities of the same color belong to the same entity chain. The visual area color corresponds to the physical color. \textbf{Gold} represents the result of gold.}
\includegraphics[width=1\textwidth]{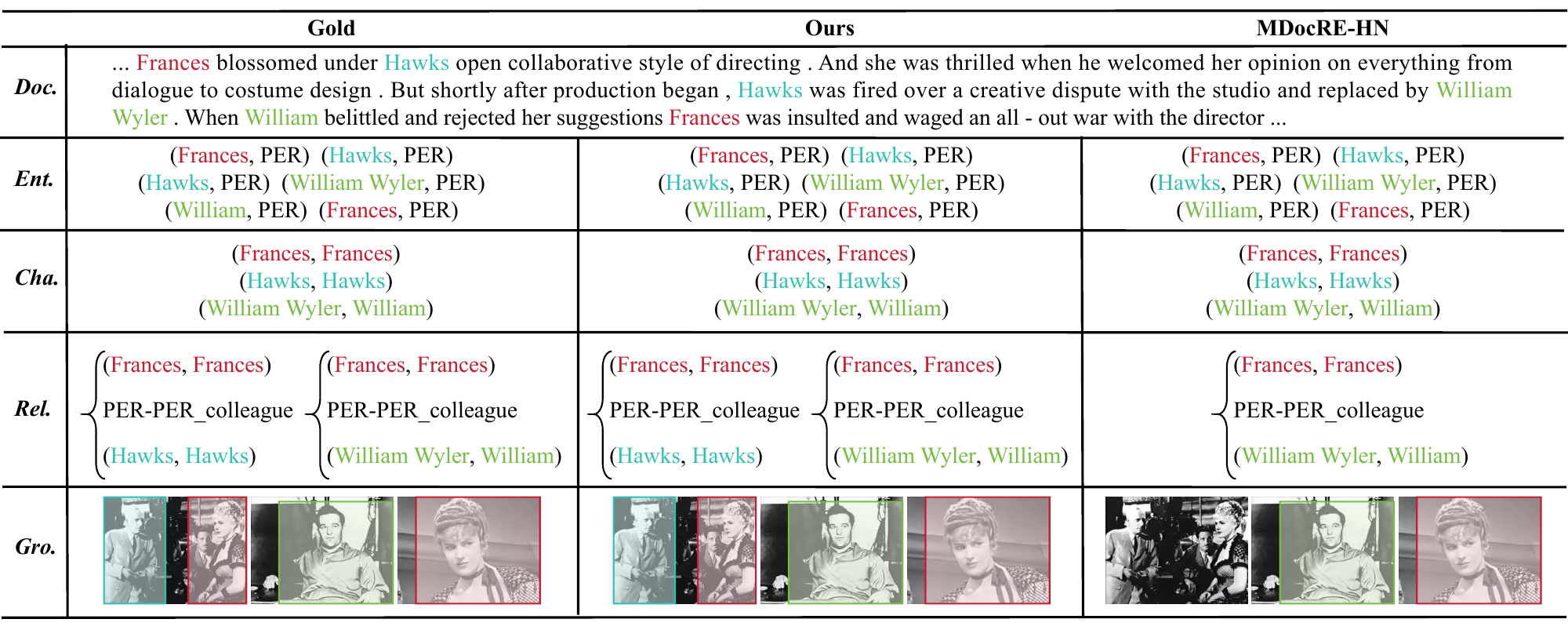}
\includegraphics[width=1\textwidth]{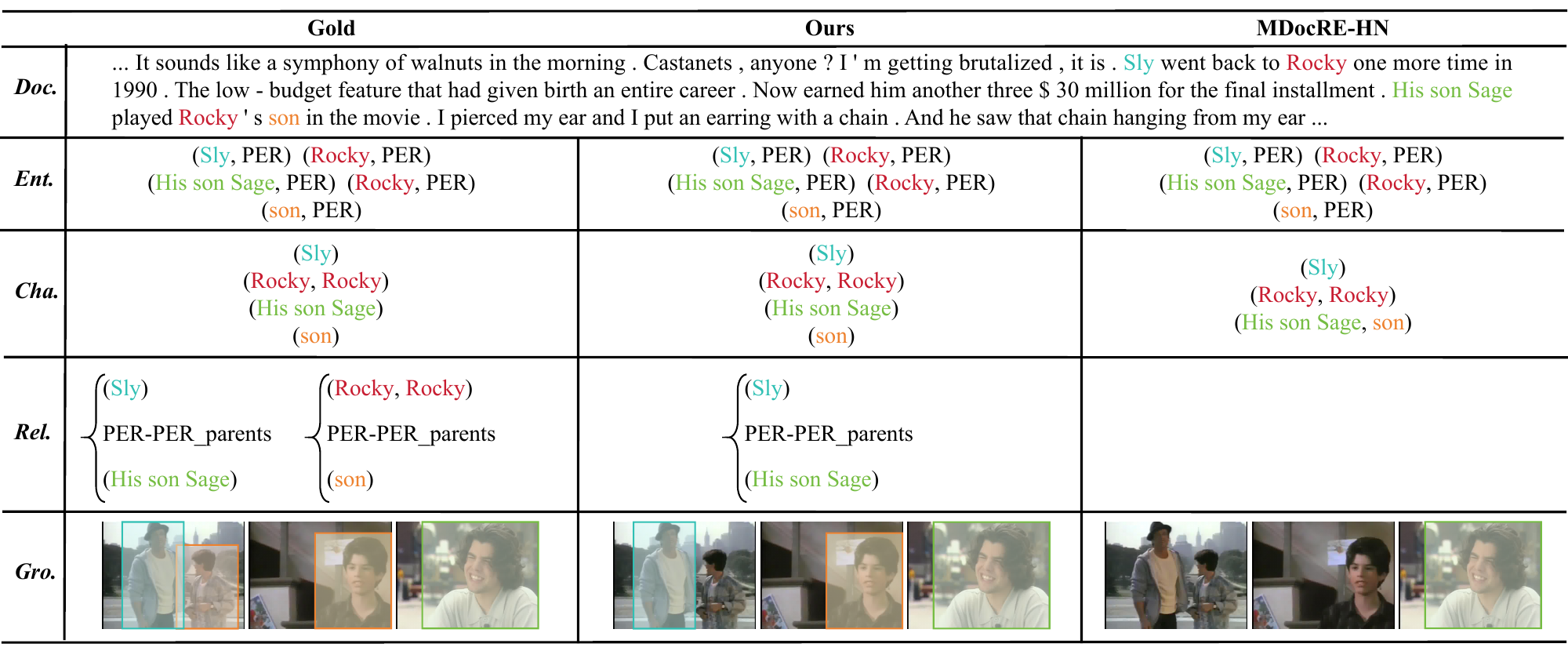}
\label{tab:case_study}
\end{table*}

\begin{table}[!t]
\caption{Error analysis of our model on four tasks. Error rate $=$ the number of prediction errors $/$ the total number of predictions. 
}
\fontsize{12}{14}\selectfont
\setlength{\tabcolsep}{0.8mm}
\setlength{\arrayrulewidth}{0.1mm}
\centering
\resizebox{0.4\textwidth}{!}{
\begin{tabular}{c c c c c c c}
\hline
\bf Task& \phantom{}& \bf Error Type& \phantom{} & \bf EN & \phantom{} & \bf ZH \\
\hline
\multirow{4}{*}{\bf Ent.}&\phantom{}&\multirow{2}{*}{Boundary Incorrect}&\phantom{}&\multirow{2}{*}{8.78} &\phantom{}&\multirow{2}{*}{12.70} \\
&\phantom{}&&\phantom{}& &\phantom{}& \\
\cdashline{2-7}
&\phantom{}&\makecell{Boundary Correct \\ but Type Incorrect}&\phantom{}&0.85 &\phantom{}&0.77 \\
\arrayrulecolor{black}\hline
\multirow{3}{*}{\bf Cha.}&\phantom{}&\makecell{Chain Contains \\Incorrect Entities}&\phantom{}&41.32 &\phantom{}&27.12 \\
\cdashline{2-7}
&\phantom{}&\makecell{No Errors in Entity Chain, \\but Entities Are Missing}&\phantom{}&10.73 &\phantom{}&5.88 \\
\arrayrulecolor{black}\hline
\multirow{3}{*}{\bf Rel.}&\phantom{}&\makecell{False Relations \\Between Chains}&\phantom{}&64.10 &\phantom{}&63.00 \\
\cdashline{2-7}
&\phantom{}&\makecell{Incorrect Relation Types\\ Between Chains}&\phantom{}&3.46&\phantom{}&3.12 \\
\arrayrulecolor{black}\hline
\multirow{4}{*}{\bf Gro.}&\phantom{}&\multirow{2}{*}{Boundary Incorrect}&\phantom{}&\multirow{2}{*}{54.15} &\phantom{}&\multirow{2}{*}{67.40} \\
&\phantom{}&&\phantom{}& &\phantom{}& \\
\cdashline{2-7}
&\phantom{}&\makecell{Boundary Correct \\but Type Incorrect}&\phantom{}&8.00 &\phantom{}&8.27 \\
\arrayrulecolor{black}\hline
\end{tabular}
}
\label{tab:Error analysis}
\end{table}

\subsection{Case Studies}

We compared the prediction results between our model and MDocRE-HN on a data sample, as shown in Table \ref{tab:case_study}. Both models are fully capable of handling the entity recognition task, where our entities are continuous and non-nested, making it the simplest compared to other tasks. In the entity chain extraction task, MDocRE-HN incorrectly identified "His son Sage" and "son" as belonging to the same entity chain in the second example. For the relation extraction task, MDocRE-HN made errors in both examples, and our model also made an error in the second example. In the visual grounding task, MDocRE-HN similarly made errors in both examples, and our model also failed in the second example. We believe there is a certain correlation here, relation extraction task is more challenging compared to entity recognition task and entity chain extraction task. As a result, when visual objects are not identified, it becomes difficult to infer relations between entity chains solely based on textual descriptions.

\subsection{Error Analysis}
We conducted an error analysis on the model. For each task, the prediction results of the model can be divided into two types. Table \ref{tab:Error analysis} shows the error analysis of our model on four tasks. For entity recognition and visual grounding tasks, their main errors come from boundary errors. The difficulty of boundary recognition is a well-known problem in previous work. \cite{bourd_error_1, bourd_error_2, vg_error}. For entity chain extraction and relation extraction tasks, their errors mainly come from false relation predictions, that is, there is no relation between  entity (chain) pair, but the model predicts that there are relations. This is because our model enumerates all entity (chain) pairs and classifies them, which can easily lead to unrelated entity (chain) pairs being predicted to have relations.

\section{Conclusion}
In this paper, we construct a multimodal, multilingual, and multitask dataset M$^{3}$D, which includes two modalities (video and text), two languages (English and Chinese), and four tasks (entity recognition, entity chain extraction, relation extraction, and visual grounding). We then design an innovative hierarchical model to solve all multimodal IE tasks and establish a The model contains denoising feature fusion modules to fully utilize and integrate multimodal information, and missing modality building modules to alleviate the problems caused by the missing modality of the model. Our work promotes the research of information extraction and multimodal content understanding. We hope that our work will attract more and more research attention in this field.

\section*{Acknowledgments}
This work is supported by the National Natural Science Foundation of China (No. 62176187).
This work is also supported by the fund of Laboratory for Advanced Computing and Intelligence Engineering.



\bibliography{references}
\bibliographystyle{IEEEtran}

\end{document}